# Long-Sequence LSTM Modeling for NBA Game Outcome Prediction Using a Novel Multi-Season Dataset

Charles Rios, Longzhen Han, Almas Baimagambetov and Nikolaos Polatidis*

School of Architecture, Technology and Engineering, University of Brighton, Brighton, U.K

*Corresponding author: N.Polatidis@Brighton.ac.uk

**Abstract**

Predicting the outcomes of professional basketball games, particularly in the National Basketball Association (NBA), has become increasingly important for coaching strategy, fan engagement, and sports betting. However, many existing prediction models struggle with concept drift, limited temporal context, and instability across seasons. To advance forecasting in this domain, we introduce a newly constructed longitudinal NBA dataset covering the 2004–05 to 2024–25 seasons and present a deep learning framework designed to model long-term performance trends. Our primary contribution is a Long Short-Term Memory (LSTM) architecture that leverages an extended sequence length of 9,840 games equivalent to eight full NBA seasons to capture evolving team dynamics and season-over-season dependencies. We compare this model against several traditional Machine Learning (ML) and Deep Learning (DL) baselines, including Logistic Regression, Random Forest, Multi-Layer Perceptron (MLP), and Convolutional Neural Network (CNN). The LSTM achieves the best performance across all metrics, with 72.35% accuracy, 73.15% precision and 76.13% AUC-ROC. These results demonstrate the importance of long-sequence temporal modeling in basketball outcome prediction and highlight the value of our new multi-season dataset for developing robust, generalizable NBA forecasting systems.

# 1. Introduction

Artificial intelligence has reshaped the ways we look at many topics, particularly sports. The use of AI in sports science has grown steadily within the recent years as shown with the technological innovation in sports leagues such as MLB, NBA, NFL. In this paper, our focus is on AI in the NBA for regular season game outcome prediction. It is evident that artificial intelligence has considerably aided in sports through multiple mediums [1-3]. Performance metrics, injury risk assessment [4], fan engagement and management are only some of the areas where AI has consistently proved its usefulness in sports. Use of AI in these areas provides major benefits towards increase in revenue for team managers and the sports league, decrease in injury for players, and better evaluation of a player's performance.

Although, exploring gaps that these tools fail to fill is necessary to utilize them efficiently. The replacement of human roles has been a major concern in all fields embracing AI, causing the widespread fear of displacement of human workers [5]. With this possible future of replacing human roles with AI, the user of AI must consider the quality and reliability of them. This brings us into another overlooked issue in AI with its need for reliability. These systems will require rigorous testing and thoughtful design, otherwise breaking down and causing problems will be more likely than solving them. Finally, one of the most pressing issues with AI use in sports is the inability to generalize well. This stems from new regulations, ever-changing nature of sports, new playstyles being adopted, and persisting attempts to maintain a fan's engagement.

Therefore, it is essential to be as efficient as possible with the data on hand as to create reliable and performant AI systems. Concept drift, which is when patterns learned in the past no longer apply effectively to the present, is a major concern in the space of AI sports models as it creates data that is no longer usable for the present's problems. This is shown through large contributors towards predictive power, like team rosters and gameplay style, significantly changing overtime. This leads to accuracy dropping as the season continues, affecting model stability and requires retraining on new data. Our new design could potentially address this issue and gain more use out of existing data. Many academics assert the idea that data from older, less recent seasons harm the model's performance. Although, this creates an environment where it's hard to develop models that can generalize and accurately predict across multiple seasons. In despite of that, we show that data from these seasons can still be taken advantage of to improve not only performance but resilience and robustness.

Our proposal of fallback or generalizable models in AI sports fills gaps in what is currently used. The replacement of human roles will not be affected by our use of AI in this paper, as our goal is to improve AI systems that have previously and currently been used. Consequently, no human jobs would be replaced in the implementation of our model as it would replace existing AI roles. Additionally, the generalizability and stability of our model combined with a specialized model would create a system that accurately and reliably forecasts game outcome. Lastly, the implementation of the LSTM architecture and a large sequence size prove to understand the patterns across seasons and not get halted by the concept drift gap that many sports models struggle with.

Thus, this article makes the following contributions:
- Introduces a new NBA basketball dataset that can be used to predict match outcomes.
- Implements a new LSTM architecture that can be applied to predict match outcomes with high accuracy.

The rest of the paper is organized as follows: section 2 describes the related works, section 3 explains the dataset and the methodology, section 4 presents the experimental evaluation and section 5 contains the conclusions.

## 2. Related Work

The use of traditional ML algorithms, statistical models, and DL architectures in research papers has been around for quite some time and has increased in popularity as of late. This section will explore how these related works support the ideas in this paper and exactly how we will fill the gaps they leave.

Recent works assert the idea that temporal dependencies improve accuracy in sports game outcome prediction. The work in [6] uses a hybrid neural network, named MambaNet, to predict the NBA playoffs. In their paper, they explain how the temporal dependencies in NBA game data are predictive and improves accuracy beyond what traditional ML can achieve without effectively capturing those relationships. Additionally, it reinforces the idea of DL models like RNN's are increasingly becoming an established tool for sports game outcome prediction. However, MambaNet was only trained on and used for playoff games, effectively ignoring any long-term patterns. Another current paper asserts similar ideas where the high dimensionality of sports statistics like time dependencies can be used to achieve higher prediction accuracies with DL architectures than those of traditional ML models [7].

Another compartment of sports game outcome prediction is the problem of recency. Concept drift, known as when the patterns learned in past data are no longer effectively useful to the present data. This is a largely known problem in sports prediction due to changes in rules, team rosters, injuries, and many other factors. Several methods have been tested to combat this, including time-decay weighting which was used in a foundational paper widely cited to maintain the idea of recent games mattering more than old one [8]. A more recent paper shows how predictive accuracy worsens with excessive number of past seasons [9]. These papers and others in the field establish a well-known issue with time-series prediction, especially with sports.

Fallback/Generalizable models in sports prediction is not very common. This may be due to problems of recency and the constantly evolving nature of the NBA in its rules and playstyle as seen with the three-point revolution with Steph Curry. However, their implications could be very beneficial towards creating sustainable and real-world applicable tools for not only basketball but all sports. One such organization, FiveThirtyEight, has created a hybrid system using ELO and RAPTOR to accurately predict NBA game outcomes [10]. How they make NBA prediction is much more than just ELO and RAPTOR, as it has been consistently improved over the years since its first major version was released in 2015. Although, their system is large and complex, utilizing multiple rating systems to come to the final prediction. The model introduced in this paper is similar in spirt to what FiveThirtyEight has accomplished with their system but simpler.

We have established that time dependencies are known to be very useful in predictions for NBA games and concept drift being a major issue to address in sports game outcome prediction. Our model aims to utilize the knowledge known in the field of sports game outcome prediction and address an issue that has plagued academics. It's nature to work as a fallback model works to aid existing architectures and tools in place with a much simpler and more efficient design.

## 3. Methodology

This section delivers the dataset in 3.1 and the proposed LSTM methodology in section 3.2

### 3.1 The dataset

Each NBA game consists of 2 teams, Home and Away. To represent both teams, each feature shown in Table 1 is duplicated in the training dataset, i.e. HOME_PTS & AWAY_PTS. This is to ensure the model's understanding of effects such as home-court advantage, which would be ignored if features were only provided for 1 team or aggregated across both. Additionally, each feature value corresponds with the statistic recorded for the given team's most recent game. This is to establish comprehension of temporal dependencies which are commonly useful for NBA game outcome prediction. This is evidenced by recent work utilizing sequential game data for RNNs, including LSTMs, to achieve higher accuracies than traditional ML models [7]. The dataset covers all NBA regular season games from 2008-2024 seasons and is fetched from nba.com/stats. While this dataset uses a diverse number of features that can show many facets of how basketball is played, it does not account for roster, mid-season trades, or free-agency signings. These unaccounted aspects of the game, if included, could have improved accuracy. The dataset can be found at https://www.kaggle.com/datasets/charlesrios/nba-advanced-metrics-2004-2024

| Feature Name | Description |
|---|---|

| Abbreviation | Description |
| --- | --- |
| WINS | Total Wins |
| LOSSES | Total Losses |
| FGM | Total Field Goals Made |
| FGA | Total Field Goals Attempted |
| FG3A | Total 3pt Field Goals Attempted |
| FG3M | Total 3pt Field Goals Made |
| FTA | Free Throws Attempted |
| FTM | Free Throws Made |
| OREB | Offensive Rebounds |
| DREB | Defensive Rebounds |
| AST | Assists |
| STL | Steals |
| BLK | Blocks |
| TO | Turnovers |
| PF | Personal Fouls |
| PTS | Points Scored |
| PLUS_MINUS | Differential between Points Scored – Points Allowed |
| OFF_RATING | Amount of Points a team produces per 100 possessions |
| DEF_RATING | Amount of Points a team allows per 100 possessions |
| AST_PCT | Percentage of a team's FGM where a player assisted on while on the court |
| AST_TOV | The ratio of the amount of assists a team made vs the number of turnovers made |
| AST_RATIO | Percentage of Team's possessions that end in an assist |
| OREB_PCT | Percentage of available offensive rebounds a team obtains |
| DREB_PCT | Percentage of available defensive rebounds a team obtains |
| TM_TOV_PCT | Percentage of possessions that end in a turnover |
| EFG_PCT | Field Goal Percentage that accounts for 3pt shots are worth more than 2pt shots |
| TS_PCT | Field Goal Percentage that incorporates free throws to measure overall scoring efficiency |
| PACE | Number of Possessions per 48 mins |
| POSS | Possessions |
| PIE | Player Impact Estimate |
| PCT_PTS_PAINT | Percentage of total points that were scored in the paint |
| PCT_PTS_FB | Percentage of total points that were scored on a fastbreak |
| PCT_PTS_OFF_TOV | Percentage of total points that were scored off a turnover |
| PCT_PTS_3PT | Percentage of total points that were scored as 3 pointers |

Table 1. Description of Features Used in Dataset

## 3.2 Proposed LSTM

We implemented an LSTM model, which is a type of Recurrent Neural Network (RNN) that is designed to handle sequential data effectively. They contain memory cells which employ a gating system that decides what information to store, forget, or send out. This allows for the LSTM to keep relevant information while discarding that which is no longer necessary. This specific structure is effective when it comes to time-series prediction like sport game outcome prediction. One of the things that is unique about LSTMs is their ability to view large amounts of data in one prediction, known as the sequence length. For example, it can view the past n games of an NBA season at a given time. For this model, the problem was structured as a sequence-to-one task where it views the past 9,840 NBA Games to make a prediction on the current game. Therefore, the sequence length is equivalent to 8 NBA regular seasons. TensorFlow and Keras were used for the implementation of the LSTM. The exact structure of the LSTM is documented in Table 2.

| *Type of Layer* | *Units / Neurons / Rate* | *Activation* | *Return Sequences* |
|---|---|---|---|
| LSTM | 200 | N/A | True |
| LSTM | 100 | N/A | True |
| LSTM | 50 | N/A | False |
| Dense | 32 | ReLu | N/A |
| Dropout | 0.3 | N/A | N/A |
| Dense | 1 | Sigmoid | N/A |

Table 2. LSTM Parameters

The proposed LSTM model is defined using the following notations where:
- $x_t$ : input vector at time t (features of the current NBA game)
- $h_{t-1}$ : previous hidden state (short-term memory from the previous game)
- $c_{t-1}$ : previous cell state (long-term memory carried across the entire sequence)
- $h_t$ : new hidden state at time t (output/short-term memory for current step)
- $c_t$ : new cell state at time t (updated long-term memory)
- $f_t$ : forget gate (decides what to discard from $c_{t-1}$)
- $i_t$ : input gate (decides what new information to add)
- $o_t$ : output gate (decides what to output as $h_t$)
- $\tilde{c}_t$ or $\hat{c}_t$ : candidate cell state (the raw new information created at time t)
- $\sigma$ : sigmoid activation function $\rightarrow \sigma(z) = \frac{1}{1+e^{-z}}$
- tanh : hyperbolic tangent activation function $\rightarrow \tanh(z) = \frac{e^z - e^{-z}}{e^z + e^{-z}}$
- $W_f, W_i, W_c, W_o$ : trainable weight matrices for the four gates
- $b_f, b_i, b_c, b_o$ : trainable bias vectors for the four gates
- $[\,h_{t-1}, x_t\,]$ : concatenation of previous hidden state and current input
- · (dot): matrix-vector multiplication

- ⊙: elementwise (Hadamard) multiplication

Then the LSTM computes six vectors using four gates as follows:

**Forget gate** which decides what to forget from the old cell state as shown in equation 1 below:

$$f_t = \sigma(W_f \cdot [h_{t-1}, x_t] + b_f) \tag{1}$$

**Input gate** which decides what new information to store as shown in equation 2 below:

$$i_t = \sigma(W_i \cdot [h_{t-1}, x_t] + b_i) \tag{2}$$

**Candidate cell update** which creates new candidate values as shown in equation 3 below:

$$\tilde{c}_t = \tanh(W_c \cdot [h_{t-1}, x_t] + b_c) \tag{3}$$

**Output gate** which decides what parts of the cell state to output as shown in equation 4 below:

$$o_t = \sigma(W_o \cdot [h_{t-1}, x_t] + b_o) \tag{4}$$

**New cell state** (long-term memory) as shown in equation 5 below:

$$c_t = f_t \odot c_{t-1} + i_t \odot \tilde{c}_t \tag{5}$$

**New hidden state** (short-term memory / output of the cell) as shown in equation 6 below:

$$h_t = o_t \odot \tanh(c_t) \tag{6}$$

Subsequently, the LSTM is adjusted to the dataset as follows:

Reads the input shape: (batch_size, 9840, features) → 9840 timesteps (past games), each with a number of statistical features and then the following steps take place:
**First LSTM layer** (200 units, return_sequences=True) For t = 1 to 9840: $h_t^{(1)} \in \mathbb{R}^{200}$ is computed using the equations above Output shape: (batch, 9840, 200)
**Second LSTM layer** (100 units, return_sequences=True) Takes the sequence of 200-dim hidden states from layer 1 as input Output shape: (batch, 9840, 100)
**Third LSTM layer** (50 units, return_sequences=False) Processes the full sequence but returns only the final hidden state $h_{9840}^{(3)} \in \mathbb{R}^{50}$ Output shape: (batch, 50)
**Dense layer** (32, ReLU) $z = W \cdot h_{9840}^{(3)} + b a = \text{ReLU}(z) = \max(0, z)$ Output: 32-dimensional vector
**Dropout layer** (0.3) Randomly sets 30% of the 32 values to zero during training (prevents overfitting)
**Dense layer** (1, Sigmoid) The final prediction layer $\hat{y} = \sigma(w \cdot a + b) = \frac{1}{1+e^{-(w \cdot a+b)}}$ Output: a single probability between 0 and 1 (e.g., probability that the home team wins)

Finally, it is worth mentioning that Deep neural networks such as LSTMs are known for the difficulty of explaining its performance. This leads to them being referred to as black boxes. A recent study shows how to combat this issue and provide more explainable AI into a field like sports [15]. The process used is known as distillation where a simpler model is trained to predict the ore complex model outputs, allowing for performance near the same level with explainable features to analyze. We believe that this process can be used to further increase the explainability in our model for those who want to adopt this architecture but value explainability over performance.

# 4. Experimental evaluation

The experimental evaluation took place on a computer using the Linux operating system and the Python programming language along with Tensorflow, Keras and Scikit learn libraries. This section contains the processing of the dataset in sub-section 4.1, the evaluation metrics in sub-section 4.2, the results in sub-section 4.3 and the discussion of the results in sub-section 4.4.

**4.1 Dataset processing**
To perform the experiments, we used the dataset described in section 3 as follows:

- The dataset includes NBA games from 2004-05 season to 2024-25 season.
- Each training sample requires a fixed sequence of 9,840 games. This sequence is useful to the LSTM for its ability to "remember" past outcomes.
- This number (9,840) comes from 1230 games (Total games scheduled to play in 1 NBA regular season) × 8 seasons.
- To build one training sample, the model takes the first 9,840 rows in the dataset (which correspond to the 2004–05 through 2011–12 seasons).
- After collecting those 9,840 games, the model predicts the outcome of the next game (the first game of the 2012–13 season).
- For the following samples, the window shifts forward by one game each time, always keeping exactly 9,840 games in the sequence.
- Although the dataset spans the 2004-05 to 2024-25 seasons, the model does not produce predictions until the 2012–13 season because everything before that season is used only to fill the initial sequence window.

**4.2 Evaluation metrics**
For the evaluation of the proposed methodology, we have used the Accuracy, Precision and AUC ROC metrics. Accuracy measures the proportion of total predictions that the model classifies correctly as shown in equation 7 where TP, TN, FP, and FN represent true positives, true negatives, false positives, and false negatives. Precision measures how many of the positive predictions made by the model are correct as shown in equation 8 where TP stands for True Positives and FP for False Positives, making it especially important when false positives carry a high cost. The Area Under the Receiver Operating Characteristic Curve (AUC-ROC) is defined in equation 9 and it evaluates how well the model separates positive and negative classes across all classification thresholds; it is computed as the integral of the true positive rate with respect to the false positive rate, and represents the probability that the model ranks a randomly chosen positive instance higher than a randomly chosen negative one and in the equation TPR is the proportion of actual positives correctly identified (TPR = TP/(TP+FN)), FPR is the proportion of actual negatives incorrectly identified as positive (FPR = FP/(FP+TN)), and d(FPR) represents an infinitesimally small change in the false positive rate used when integrating the ROC curve.

$$Accuracy = \frac{TP + TN}{TP + TN + FP + FN} \tag{7}$$

$$Precision = \frac{TP}{TP + FP} \tag{8}$$

$$AUC\ ROC = \int_0^1 TPR(FPR)\, d(FPR) \tag{9}$$

## 4.3 Experimental results

For comparison purposes we have used the Logistic Regression, Random Forest, Multi-Layer Perceptron and CNN classifiers as explained below. Moreover, we compared the proposed methodology against three recent works as shown later.

**Logistic Regression**
We started with Logistic Regression (LR) to create a baseline to test against the more complex models as it is commonly treated as a starting point for classification problems, especially NBA game outcome prediction [11-13]. LR estimates the probability of a binary outcome, in this case (Home Win/Away Win), and learns by using a linear combination of features with a sigmoid activation function. Scikit-Learn was used for the implementation of the LR model and L2 regularization was applied to prevent overfitting and increase generalization across the dataset. Moreover, the L2 regularization penalty parameter was set at 1.0.

**Random Forest Classifier**
We then attempted to use a Random Forest Classifier (RF) to see how accuracies would compare. RFC is a ML algorithm that builds multiple decision trees during training and merging their predictions. Each one of the trees is trained on a random subset from the training data. This is done by a statistical technique known as bootstrapping where samples from the dataset are repeatedly chosen at random from a subset of samples. Once a sample is chosen, it is put back into the subset of samples before the next pick. In the context of Random Forests, this guarantees that each tree in the forest views a different dataset, which in turn aids in preventing overfitting and increasing diversity. At each split in the tree, only a random subset of features is considered. Scikit-Learn was used in the implementation of the Random Forest Classifier. The randomness of RFCs is what leads them to be less prone to overfitting and more generalizable. Tree-based models like RFCs have been becoming a very reasonable choice in architecture for sports game prediction. Compared to ANNs, they need much less data, easier to implement, and less prone to overfitting. They are very useful when data is limited, which can be the case in NBA game outcome forecasting, and perform well without careful tuning [14].

**Multi-Layer Perceptron**
Now, we move onto DL architectures. The MLP was used to increase accuracy with its increased complexity compared to Random Forests and Logistic Regression. An MLP is a form of an artificial neural network (ANN) that contains 1 input layer, 1 or more hidden layers, and 1 output layer. It takes in inputs from the input layer and passes the inputs through the hidden layers that uses neurons to transform them using biases, weights and activation functions. Finally, the output layer gives the prediction. Within each neuron, a weighted sum of the inputs is computed, and a bias is added. The sum of inputs is then passed through a nonlinear activation function. The output of one layer becomes the input of the next. Through this process, the MLP adapts and learns to link inputs to outputs by minimizing a given loss function. TensorFlow and Keras were used for the implementation of the MLP. For the exact architecture of MLP refer to Table 3. The loss function is binary cross entropy as this is standard for a binary classification problem and the optimizer used is Adam.

| *Type of Layer* | *Neurons / Rate* | *Activation* |
|---|---|---|
| Dense | 128 | Relu |
| BatchNormalization | N/A | N/A |
| Dropout | 0.3 | N/A |
| Dense | 64 | Relu |
| Dropout | 0.3 | N/A |
| Dense | 32 | Relu |
| Dense | 1 | Sigmoid |

**Table 3**. Multi-Layer Perceptron Parameters

**Convolutional Neural Network**

Entering the realm of more complex architectures, we tested against a CNN. CNNs have convolutional layers that use filters that slide over the input to identify patterns. Pooling layers can be added to reduce the size and complexity by summarizing nearby values to prevent overfitting. The last connected layers combine the extracted features to make the final prediction. We used L2 regularization and 3 Pooling layers to prevent overfitting and increase generalization. TensorFlow and Keras were used for the implementation of the CNN. Binary cross entropy was used as the loss function and Adam is the optimizer. The exact structure of the CNN is documented in Table 4.

| *Type of Layer* | *Neurons* | *Kernel Size* | *Pool Size* | *Activation* | *Kernel Regularizer Value* |
|---|---|---|---|---|---|
| Conv2D | 32 | (1,2) | N/A | ReLu | 0.001 |
| MaxPooling2D | N/A | N/A | (1,2) | N/A | N/A |
| Conv2D | 64 | (1,2) | N/A | ReLu | 0.001 |
| MaxPooling2D | N/A | N/A | (1,2) | N/A | N/A |
| Conv2D | 128 | (1,2) | N/A | ReLu | 0.001 |
| MaxPooling2D | N/A | N/A | (1,2) | N/A | N/A |
| Flatten | N/A | N/A | N/A | N/A | N/A |
| Dense | 128 | N/A | N/A | ReLu | 0.001 |
| Dense | 1 | N/A | N/A | Sigmoid | N/A |

**Table 4.** CNN Parameters

All models were evaluated on Accuracy, Precision and AUC-ROC effectively as shown by the results in figures 1, 2 and 3 respectively. AUC-ROC was specifically utilized to assess each model's ability to distinguish between classes. This metric is important for any classification model, but especially for sports game outcome prediction as illustrated by Vegas betting lines which calculate win probabilities for sports games. The forecasts given by these betting lines are roughly 55% accurate which is significantly lower than out model's accuracy of 72.35% [16].

Reliability on the model's positive classification (Home Win) is analyzed by the precision metric. False positives can be costly with NBA game outcome prediction for NBA teams where 1 or 2 false positives can derail a strategy towards making the NBA playoffs. Additionally, sportsbooks using these types of models need to be able to rely on positive classifications to create accurate odds or their revenue will see a downtrend.

The Logistic Regression model attained a respectable accuracy of 70.12%. This creates a simple baseline to compare to other models. AUC-ROC and Precision also reached moderate levels with 69.85% and 70.66% respectively. Although, since logistic regression underperformed compared to the DL models, with exception to the CNN, this shows that the linear relationships are inadequate alone for fully capturing the patterns in the data.

The Random Forest performed similarly to the baseline with a slightly increased accuracy and AUC-ROC score but decreased precision. This means that while the random forest can capture patterns in the data slightly better than the baseline, but it needed to sacrifice reliability on its predictions to accomplish that. This may reflect a precision-recall trade off, which might be due to the model being more sensitive towards the positive class (Home Win). However, the random forest did provide more information on the data then simply accuracy. This model contains feature importances after being fitted on the data which shows how much each individual feature contributed to the final prediction. The Top 3 metrics that are most predictive for NBA game outcome prediction are LOSSES, WINS, and PIE.

While being known for their complexity and strength in identifying patterns, MLPs are the simplest form of DL models. Here, there is a larger jump in accuracy seen with the Logistic Regression Vs. Random Forest as well as a much larger increase in AUC and Precision. This demonstrates how the non-linear patterns in the data are more significant for this task than the linear patterns captured with the baseline. Also, the complexity of the model is aiding with capturing the complex patterns of the data.

CNNs are commonly used for visual or graph-like data, but they can be leveraged to work with tabular data. They have seen some use in NBA game outcome prediction due to their complexity and ability to capture spatial patterns effectively. There is a decrease in both accuracy and precision but an increase in AUC-ROC as seen in the results chart, indicating that the CNN is better at creating win probabilities. The model can understand probability distributions more proficiently but underperforms when it comes to actual predictions and classifying positives. To combat this, threshold optimization and probability calibration might have a positive effect on its overall performance and while CNN's can be used for tabular data, this is not their area of expertise and are commonly leveraged with other models (i.e. hybrid architectures). Shot charts are one example of where CNNs can shine in sports game prediction especially when used as part of an intelligent system recognizing both spatial and temporal patterns [2]. Those patterns are well known to be predictive and present in sports game data. Finally, the LSTM model performed exceptionally well compared to the baselines. It achieved the highest scores across all metrics of any of the models. The LSTM's performance suggests that the temporal dependencies captured are the most effective for this task and contribute to predictive accuracy.

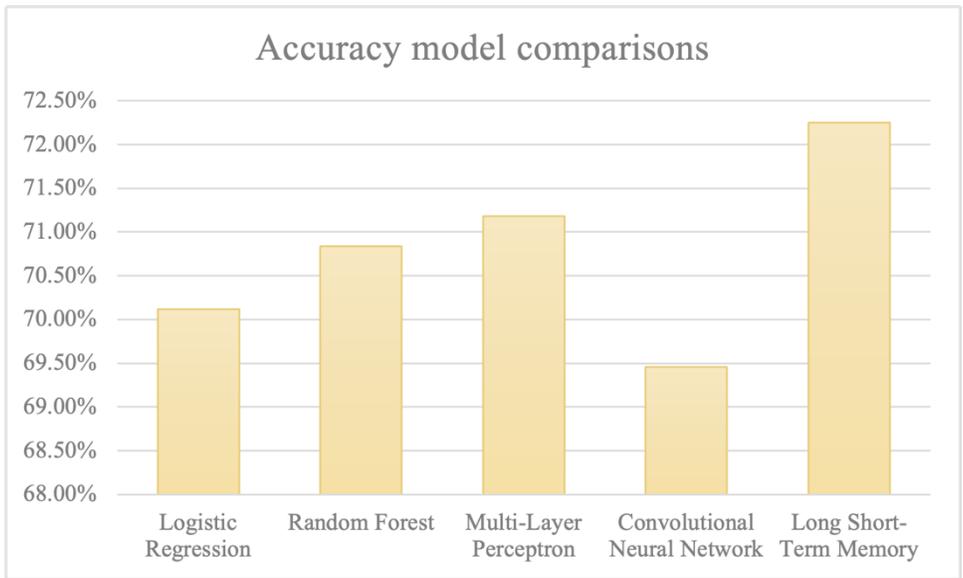
**Figure 1**. Accuracy results Comparisons

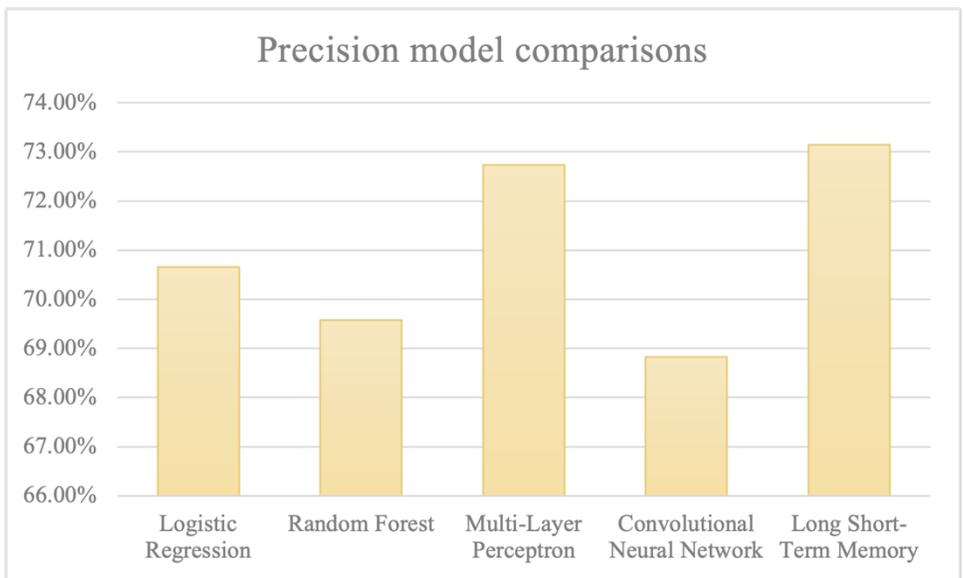
**Figure 2**. Precision results Comparisons

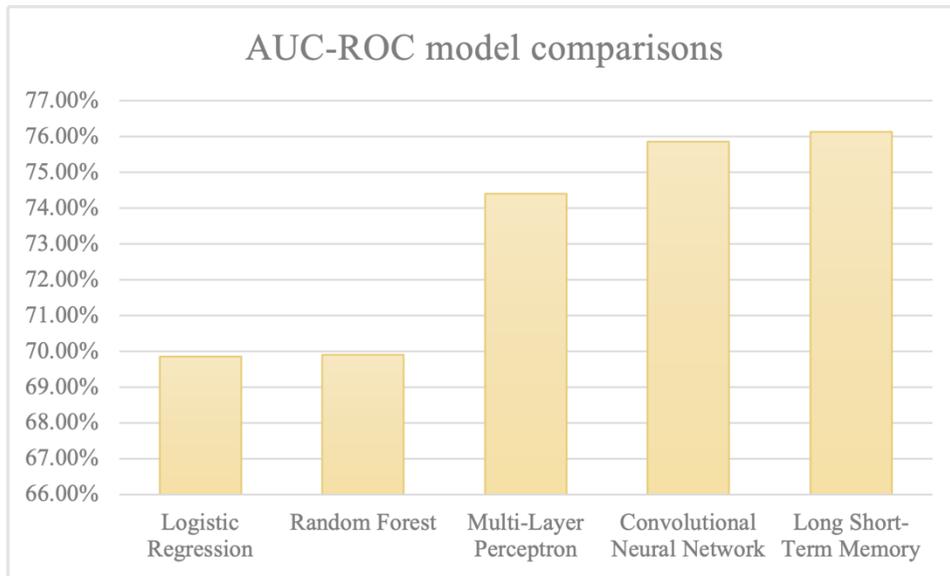

**Figure 3**. AUC-ROC results Comparisons

In addition to the previous comparisons, we compared our proposed methodology against three state-of-the-art methods as shown below and in figure 4.
- **Predicting Results for Professional Basketball Using NBA API Data** by Perricone, J., Shaw, S., & Swiechowicz, J. (2016) Ref [12].
- **Predicting Season Outcomes for the NBA** by Teno, G. D. S., Wang, C., Carlsson, N. & Lambrix, P. (2022) Ref [17].
- **Enhancing Basketball Game Outcome Prediction through Fused Graph Convolutional Networks and Random Forest Algorithm** by Zhao, K., Du, C. & Tan, G. (2023) Ref [18].

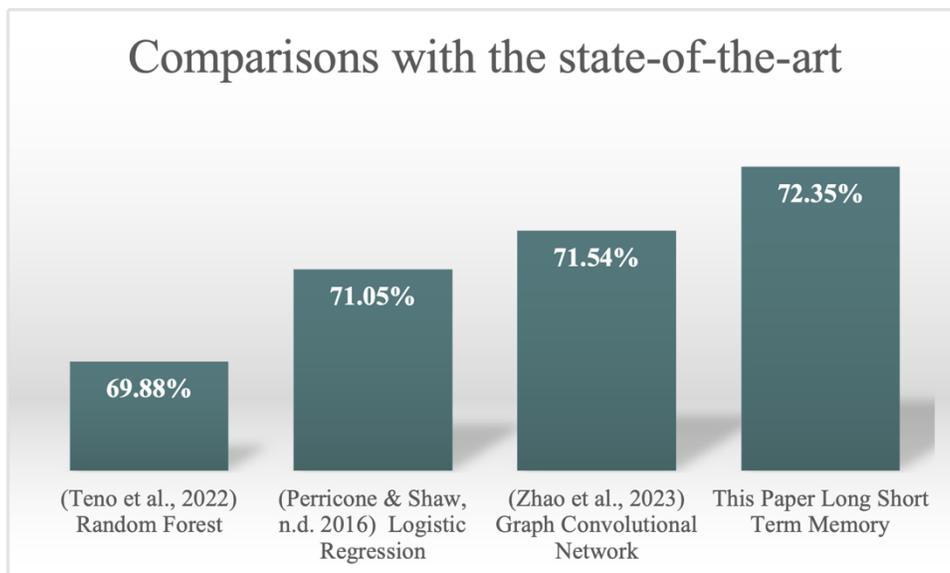

**Figure 4.** Comparison with previous methods based on the accuracy metric

## 4.4 Discussion

The LSTM model performed the best out of the 5 models that were tested. This suggests that the temporal dependencies in the data were much better at the task then the linear relationships learned through logistic regression and the non-linear relationships learned with the MLP and CNN. While RFCs can handle sequential data and temporal dependencies, they are not as optimized and tailored for that data as the LSTMs are. The performance of each model was as expected except for the CNN. We had expected the CNN to achieve a higher accuracy score than the MLP and especially the Logistic Regression and Random Forest models, however, its only improved metric was AUC-ROC. Due to the structure and goals of CNN architecture, tabular data is not usually where the neural networks shine. We believe this is why it did not perform better. Tabular data can be transformed in a way such that CNNs can view them in channels and treat them as images, however, the results of this study imply that time sequences are more useful in predicting NBA games than data transformed for CNNs. The domain that CNNs excel in is image detection and spatial pattern recognition which utilizes image or graph data. Possibly, if image data, such as shot chart plotting where each team's made and missed shots on the court, were used, then the CNN might have performed reasonably better.

Another interesting observation of the results is in the feature importances of the Random Forest model. The NBA has gone through a massive change over the years, and this is largely due to the three-point revolution and the recognition of the strength of 3 pointers. This notion is statistically supported by a study showing the long-term trends of 3pt and 2pt shots over 40 consecutive seasons of the NBA [19]. Teams have shifted coaching styles to include more 3pt shots and acquire players who are skilled in making 3 pointers as a winning strategy, i.e. Golden State Warriors in the late 2010s and the rise of the shooter player archetype. For this reason, we had expected FG3A and FG3M would be some of the highest ranked features by importance. Yet, those features ranked near the bottom while other features like WINS, LOSSES, PIE, and even PCT_PTS_PTS_PAINT were much more important towards the final prediction. While Zajac et al, demonstrated how the 3pt shot has been trending upwards, our study shows that the 3pt shot does not contribute heavily to overall interseason game outcome, contradicting our initial hypothesis. This may be occurring due to the model's generality across seasons not being bias towards large changes, like ignoring other statistics that are still predictive. The LSTM outperforming the other models that do not have architecture specifically for dealing with time-series data shows how important the time dependencies are in NBA game outcome prediction, which supports the notion of time dependencies' strong predictive power in sports data [7].

Our model also outperformed several papers in recent years (Figure 2), showcasing the gap in the space that this new implementation of LSTM models could fill. However, we propose our model as being used as a generalizable fallback, that aims to support single-season models in reliability and precision. We believe that the large sequence history allows for the LSTM to learn generalizable patterns across teams and seasons. With our model as a fallback, sports prediction systems would not only be more robust but could also increase accuracy when stacking with other models that are trained specifically for the current season. In theory, this could help academics and professional in being more efficient with increasingly large amounts of data. Being able to make use of data that was previously believed to be unusable would not only increase efficiency in the field but possibly open the door for new ideas to flourish.

There are still limitations to consider when applying this structure that existing frameworks may not have to. A large amount of data is needed to put this into effect. In terms of NBA regular seasons, the sequence size alone is 8 and to adequately train and test a DL architecture like LSTM requires large train and test sets. This is especially true for the ability to generalize across seasons, meaning the train and test sets must span across multiple seasons. Therefore, many seasons of NBA game data is needed to incorporate this setup. Another caveat is that our study only shows the results of predicting NBA regular season games, which does not include the post or pre-season. Therefore, ability in predicting NBA playoffs or summer league games is unknown. Our hypothesis would be that there is not enough data in the post or pre-season to utilize this structure as they contain significantly less games.

## 5. Conclusion

In conclusion, this study has showcased how models can generalize across multiple seasons for sports game outcome prediction and be majorly beneficial to professionals in the field. With the LSTM's tailored architecture for dealing with time series data, its performance demonstrates that it can handle large amounts of data and make accurate predictions. This may lead to its consideration for creating generalizable fallback models for NBA Game Outcome prediction. The precision and AUC-ROC score provide evidence to how well the model can distinguish between classes and how reliable its predictions are. This is essential for multiple sectors of the sports industries such as coaches, trainers, players, analysts, sportsbooks, and bettors.

 Before getting into the implications of this proposal, let's explore where it falls short. First, this proposed model is limited by its large sequence size. Using such a large lookback is what allows the model to generalize across multiple seasons and identify key patterns in game outcome that affect all teams. Handling such a large amount of data can be tasking due to the need of sufficient processing power and high storage requirements. This also contributes to the difficulty of transitioning this framework towards real-time predictions. If one was to use this architecture for real-time live predictions, they would need to understand that the model might not be able to keep up with real-time speed while processing a large sequence size. Second, its use in non-regular season games, such as post-season or pre-season games, is expected to be inadequate. The NBA playoffs and NBA Summer League are fundamentally different than the regular season. For example, teams during playoffs utilize their best players and strategies to win the championship, therefore, the intensity and importance of the game is much higher, leading to differences in playstyle and increased risk of injury which are not present in the regular season. Additionally, there are drastically less post-season and pre-season game than the regular season which creates much less data to train on, increasing the risk of overfitting with a DL model like a LSTM used in this paper. Finally, this model would be much more effectively utilized as a fallback model as suggested in this proposal. Providing an educated prior to a more specialized model, our model will use generalized patterns across seasons that the specialized model cannot consider. The specialized model fine-tuning that prior with current season patterns will result in more accurate, reliable, and stable predictions. The limitation stems from handling these models which in turn uses more resources.

 Let's dive deeper into the contributions that this type of model can have for the industry. Coaches would be able to make more informed decisions [20-22]. In a situation where a given team was predicted to lose, coaches could rest star players to prevent over-exhaustion and injuries, test new team-strategies in a real-game setting, and evaluate bench player performance to identify ones with potential for stardom. Players would benefit from increased rest days and less risk of injuries. In recent years, the NBA and its player-base have had issues with load management, where a healthy player is held from playing in a game for increased resting and decrease in long-term injury risk. To combat this, in 2023, the NBA implemented a new Player Resting Policy (PRP) which made it more difficult for teams to rest star players [23]. Effectively, decreasing the number of rest opportunities that are available. This model allows coaches to better utilize the limited rest days they have for their star players.

Additionally, the contributions span towards the sports betting sector and aids both sportsbooks and individual bettors in deciding how to act [24-26]. For sportsbooks, the use of AI in calculating odds is already in place. However, generalizable multi-season fallback models could be the next step in producing more reliable information for calculating odds, especially when ensembled or stacked with a season or team-specific model. This would lead to better calibrated odds and increased profits for the sportsbook. For individual bettors, use of this model allows for consideration of statistics and features that they cannot without arduous and time-consuming analysis. Which will in turn, provide for better informed decision-making and possibly new strategies for increasing their return on investment.

This paper also confirms the notion of temporal dependencies being very useful in sports game outcome prediction. In the future, to improve the model further, we would include team rosters, game lineups, injuries, trades, and even off-season moves made by teams. There are many facets of the NBA and sports in general that come from off the court which could have increased accuracy in this case. Nonetheless, this idea is not limited to only the NBA. Its design could potentially be used in other sports such as baseball, football, and many others. This study highlights not only the potential of a future in sports with AI but can create new and interesting ways to watch the sports that give people around the world such enjoyment.

**Declarations**

**Data availability statement** We have made the dataset publicly available at https://www.kaggle.com/datasets/charlesrios/nba-advanced-metrics-2004-2024

**Competing interests** The authors declare that there are no competing interests.

**Funding** The authors did not receive any funding.

**Acknowledgements**
For the purpose of open access, the authors have applied a Creative Commons Attribution (CC BY) licence to any Author Accepted Manuscript version arising from this submission